\documentclass[10pt,twocolumn,letterpaper]{article}

\usepackage{iccv}              

\usepackage{stfloats} 

\usepackage[most]{tcolorbox}

\newtcolorbox{disclaimerbox}{
    colback=gray!10,     
    colframe=gray!40,    
    boxrule=0.5pt,       
    arc=4pt,             
    auto outer arc,
    boxsep=5pt,
    left=6pt,
    right=6pt,
    top=4pt,
    bottom=4pt,
    enhanced jigsaw
}

\definecolor{iccvblue}{rgb}{0.21,0.49,0.74}
\usepackage[pagebackref,breaklinks,colorlinks,allcolors=iccvblue]{hyperref}

\def\paperID{*****} 
\def\confName{ICCV}
\def\confYear{2025}

\title{Efficient Perceptual Image Super Resolution: AIM 2025 Study and Benchmark}

\author{
Bruno Longarela~$^\diamond$ \hspace{-2mm}
\thanks{B.~Longarela (\texttt{brulon@cidaut.es}, Cidaut AI, Spain) and M.~V.~Conde (\texttt{marcos.conde@uni-wuerzburg.de}, Cidaut AI and University of W\"urzburg) are the corresponding authors.\\ 
AIM 2025 webpage: \href{https://cvlai.net/aim/2025}{https://cvlai.net/aim/2025}.\\
Code: \href{https://github.com/brulonga/AIM-2025-Efficient-Perceptual-SR-Challenge}{https://github.com/brulonga/AIM-2025-EPSR-Challenge}\\
PSR4K: \href{https://drive.google.com/file/d/1xKMya4Oqqud9__O5YyLis6QJLZJHYm6d/view?usp=sharing}{https://drive.google.com/file/PSR4K-LR.tar.gz}
}
~~~~~
Marcos V. Conde~$^{\diamond\dagger}$ ~~~~~
Alvaro Garcia~$^\diamond$ ~~~~~ 
Radu Timofte$^\dagger$ \\ \\
{$^\diamond$ Cidaut AI \quad $^\dagger$ University of W\"urzburg, Computer Vision Lab}
}
\begin{document}
\maketitle

\begin{abstract}
This paper presents a comprehensive study and benchmark on Efficient Perceptual Super-Resolution (EPSR). While significant progress has been made in efficient PSNR-oriented super resolution, approaches focusing on perceptual quality metrics remain relatively inefficient. Motivated by this gap, we aim to replicate or improve the perceptual results of Real-ESRGAN while meeting strict efficiency constraints: a maximum of 5M parameters and 2000 GFLOPs, calculated for an input size of 960 × 540 pixels. The proposed solutions were evaluated on a novel dataset consisting of 500 test images of 4K resolution, each degraded using multiple degradation types, without providing the original high-quality counterparts. This design aims to reflect realistic deployment conditions and serves as a diverse and challenging benchmark. The top-performing approach manages to outperform Real-ESRGAN across all benchmark datasets, demonstrating the potential of efficient methods in the perceptual domain. This paper establishes the modern baselines for efficient perceptual super resolution.
\end{abstract}    
\section{Introduction}
\label{sec:intro}

Single-image super-resolution (SR) aims to reconstruct a high-resolution image from a low-resolution input, which is a fundamentally ill-posed inverse problem. Traditionally, bicubic down-sampling has been the standard degradation model due to its simplicity and reproducibility. Models trained solely on bicubic degradation, however, perform poorly when confronted with complex real-world degradations such as noise, JPEG compression artifacts, and various types of blur~\cite{conde2022swin2sr}.

Optimizing exclusively for distortion metrics like PSNR or SSIM tends to produce overly smooth results due to regression to the mean, resulting in outputs that lack high-frequency details and perceptual quality. This phenomenon is theoretically supported by the perception–distortion tradeoff, which establishes that improving both distortion and perceptual quality simultaneously is fundamentally limited \cite{Blau_2018, Blau_2018_ECCV_Workshops}. Consequently, approaches incorporating perceptual losses have been shown to significantly improve the naturalness of generated images, although often at the expense of traditional metrics (PSNR or SSIM) \cite{johnson2016perceptuallossesrealtimestyle}.

State-of-the-art methods in perceptual super-resolution have traditionally relied on generative adversarial networks (GANs), with notable models such as SRGAN \cite{ledig2017photorealisticsingleimagesuperresolution}, ESRGAN \cite{wang2018esrganenhancedsuperresolutiongenerative}, Real-ESRGAN \cite{wang2021realesrgantrainingrealworldblind}, and BSRGAN \cite{zhang2021designingpracticaldegradationmodel} demonstrating visual quality improvements. Recently, diffusion-based models have emerged as strong alternatives. Notably, SR3 \cite{saharia2021imagesuperresolutioniterativerefinement} employs denoising diffusion probabilistic models (DDPMs) to iteratively refine images, while latent diffusion models (LDMs) \cite{rombach2022highresolutionimagesynthesislatent} improve efficiency by operating in a compressed latent space. These score-based generative methods achieve superior perceptual quality but remain computationally demanding, limiting their suitability for real-time or resource-constrained applications.

While GAN-based methods are generally more efficient than diffusion models, their computational demands still pose challenges for deployment on mobile and edge devices, where low latency and limited hardware resources are critical. In contrast, PSNR-oriented super-resolution methods have benefited from extensive research and optimization \cite{Timofte_2017_CVPR_Workshops, li2022ntire2022challengeefficient, inproceedings, ren2024ninthntire2024efficient, ren2025tenthntire2025efficient, Zamfir_2023_CVPR, Conde_2023_CVPR}, successfully pushing the boundaries of efficiency and performance.

Despite these advances, the development and benchmarking of efficient perceptual super-resolution methods remain largely unexplored. 
This study and benchmark aims to bridge the divide between visual quality and efficiency.

\paragraph{Related Challenges}
This challenge is one of the AIM 2025~\footnote{\url{https://www.cvlai.net/aim/2025/}} workshop associated challenges on: high FPS non-uniform motion deblurring~\cite{aim2025highfps}, rip current segmentation~\cite{aim2025ripseg}, inverse tone mapping~\cite{aim2025tone}, robust offline video super-resolution~\cite{aim2025videoSR}, low-light raw video denoising~\cite{aim2025videodenoising}, screen-content video quality assessment~\cite{aim2025scvqa}, real-world raw denoising~\cite{aim2025rawdenoising}, 
perceptual image super-resolution~\cite{aim2025perceptual}, 
efficient real-world deblurring~\cite{aim2025efficientdeblurring}, 4K super-resolution on mobile NPUs~\cite{aim20254ksr}, efficient denoising on smartphone GPUs~\cite{aim2025efficientdenoising}, efficient learned ISP on mobile GPUs~\cite{aim2025efficientISP}, and stable diffusion for on-device inference~\cite{aim2025sd}. Descriptions of the datasets, methods, and results can be found in the corresponding challenge reports.

\section{Efficient Perceptual Super Resolution Benchmark}
\label{sec:challenge}

The goals of the proposed study and benchmark are: (i) to improve the state of the art in perceptual super-resolution by encouraging the development of models that balance high visual quality with computational efficiency, (ii) to provide a standardized benchmark and platform where diverse approaches can be rigorously compared under consistent efficiency and quality constraints, and (iii) to foster collaboration and knowledge exchange between academic researchers and industry professionals, accelerating progress toward deployable, real-time perceptual SR solutions.
This section presents an in-depth description of the challenge.

\begin{figure}[t]
    \centering
    \begin{subfigure}[b]{0.57\columnwidth}
        \includegraphics[width=\textwidth]{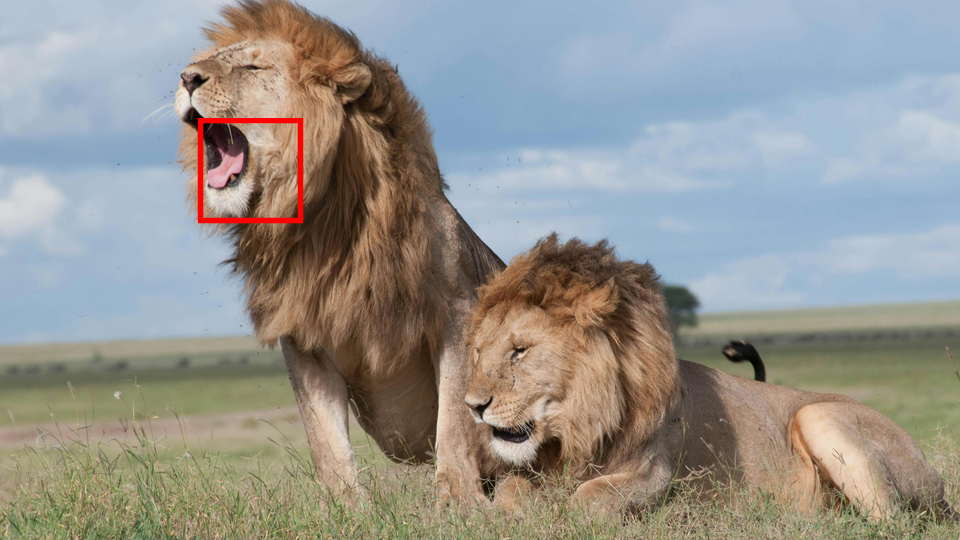}
    \end{subfigure}
    \begin{subfigure}[b]{0.32\columnwidth}
        \includegraphics[width=\textwidth]{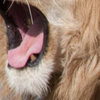}
    \end{subfigure}
    \begin{subfigure}[b]{0.57\columnwidth}
        \includegraphics[width=\textwidth]{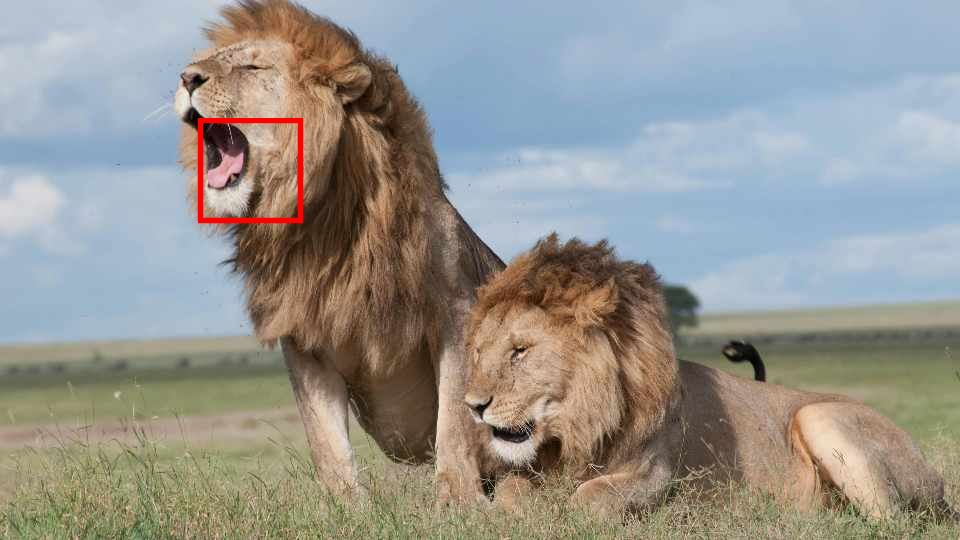}
    \end{subfigure}
    \begin{subfigure}[b]{0.32\columnwidth}
        \includegraphics[width=\textwidth]{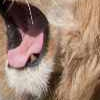}
    \end{subfigure}
        \begin{subfigure}[b]{0.57\columnwidth}
        \includegraphics[width=\textwidth]{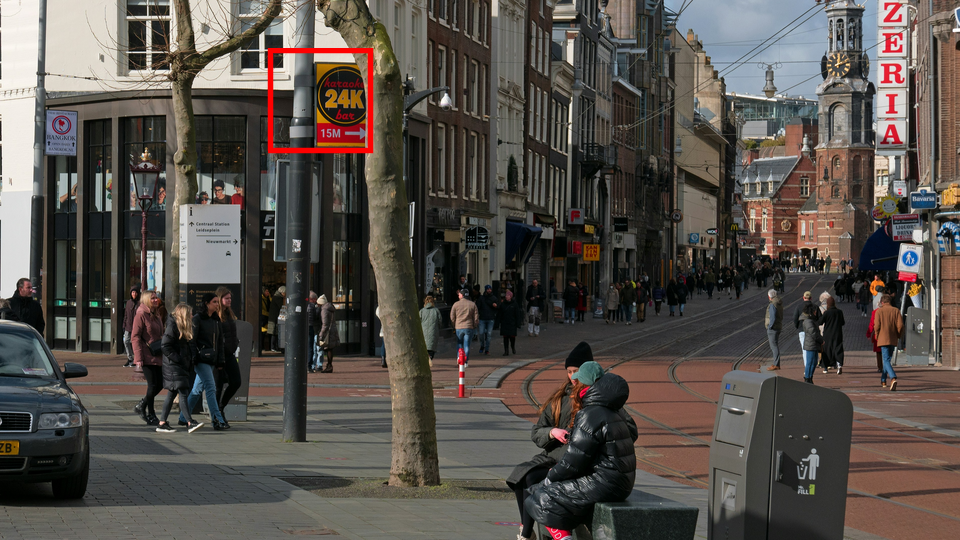}
    \end{subfigure}
    \begin{subfigure}[b]{0.32\columnwidth}
        \includegraphics[width=\textwidth]{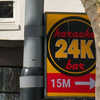}
    \end{subfigure}
    \begin{subfigure}[b]{0.57\columnwidth}
        \includegraphics[width=\textwidth]{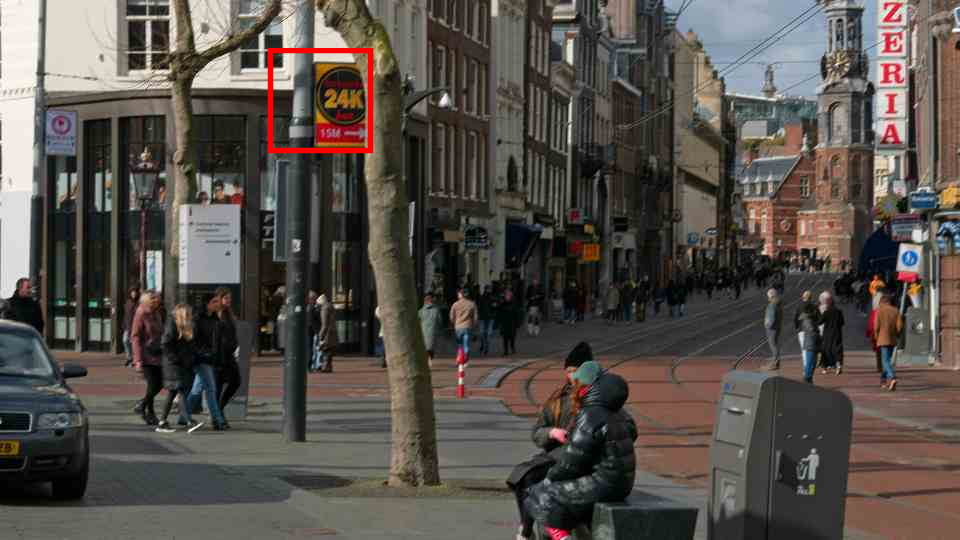}
    \end{subfigure}
    \begin{subfigure}[b]{0.32\columnwidth}
        \includegraphics[width=\textwidth]{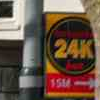}
    \end{subfigure}
    \caption{We show two samples of high-resolution (HR) and low-resolution (LR) images. For memory and layout considerations in this document, the HR images have been down-scaled to match the spatial dimensions of the LR images. Consequently, the visual differences between HR and LR examples may appear less pronounced in this figure. It should be noted, however, that the original LR images in the dataset are relatively large in resolution.
    }
    \label{fig:dataset}
\end{figure}

\subsection{Datasets}
\label{subsec:datasets}

\paragraph{Training Datasets} In this challenge, participants were free to choose their training datasets. The most commonly used datasets among participants were DIV2K \cite{8014884}, Flickr2K \cite{lim2017enhanced}, LSDIR \cite{Li_2023_CVPR}, and OST \cite{wang2018recoveringrealistictextureimage} (see training details in \Cref{sec:methods}).

\begin{itemize}
\item DIV2K: 800 high-quality images at 1K-2K resolution.
\item Flickr2K: 2,650 images at 1K-2K resolution, typically used alongside DIV2K for super-resolution training.
\item LSDIR: 86,991 high-quality images at 1K-2K resolution.
\item OutdoorSceneTrain (OST): 10,324 images at 1K or 2K resolution. Originally introduced as a segmentation dataset by Wang et al. (2017) \cite{wang2018recoveringrealistictextureimage}, later repurposed for super-resolution in Real-ESRGAN \cite{wang2021realesrgantrainingrealworldblind}.
\end{itemize}

The numbers above refer to the training splits. Some datasets include official validation sets (e.g., DIV2K with 100 images and LSDIR with 1,000 images), while others do not. For instance, Flickr2K is typically used only for training, and OST contains only a test split of 300 images.

The degradation pipelines applied during training were not fixed, each method uses slightly different variants based on Real-ESRGAN \cite{wang2021realesrgantrainingrealworldblind} degradation pipeline. The down-scaling factor was set to ×4. 

\paragraph{Testing Datasets} We evaluate diverse methods using our novel dataset \textbf{PSR4K} and no-reference image quality assessment (NR-IQA). This dataset consists of 500 low-resolution (LR) images at 960×540 pixels, grouped into ten categories: animals, architecture, art, food, nature, objects, portraits, sports, text, and urban scenes. For each category, five different degradations were applied, involving various down-sampling methods, blurs, and JPEG compressions. The exact degradation pipeline remains private to ensure the integrity of the benchmark. The chosen input resolution produces ultra-high-definition (UHD) outputs (3840×2160 pixels) at ×4 scaling. You can see some examples of our dataset and degradations in \Cref{fig:dataset}.

Additionally, methods were tested on existing perceptual SR NR-IQA benchmark datasets:

\begin{itemize}
\item PIPAL validation dataset: 1,000 images (288×288 pixels) with 40 types of degradations, including GAN-based degradations. \cite{gu2020pipallargescaleimagequality}.
\item DIV2K-LSDIR validation dataset: the 100 DIV2K validation images combined with 100 LSDIR validation images, degraded using only bicubic down-sampling \cite{8014884, Li_2023_CVPR}.
\item RealSR validation dataset which consists of 100 images exhibiting real-world degradations commonly used in perceptual super-resolution benchmarks \cite{cai2019realworldsingleimagesuperresolution}.
\item RealSRSet: 20 images with complex degradations. \cite{zhang2021designingpracticaldegradationmodel}.
\item Real47: 47 images with complex degradations.\cite{lin2024diffbirblindimagerestoration}.
\end{itemize}


\subsection{Preliminaries}
\label{subsec:rules}

\paragraph{\textbf{Baseline}}
The Real Enhanced Super-Resolution GAN (Real-ESRGAN) \cite{wang2021realesrgantrainingrealworldblind} is adopted as the baseline. It models complex real-world degradations using a second-order pipeline that repeatedly applies blur, noise, resizing, and compression, including sinc filtering. The generator uses ESRGAN’s residual-in-residual dense blocks (RRDB) with pixel-unshuffle for efficiency, while a U-Net discriminator with spectral normalization stabilizes GAN training and improves per-pixel feedback. Training is performed in two stages: PSNR-oriented pretraining with L1 loss, followed by fine-tuning with L1, perceptual, and adversarial losses. Ground-truth sharpening is applied during training to balance sharpness and artifact suppression.

\paragraph{\textbf{Efficiency constraints}}
The efficiency limits are fixed at 5M parameters ($\approx30\%$ of Real-ESRGAN) and 2000 GFLOPs ($\approx22\%$ of Real-ESRGAN), measured for an input size of $960 \times 540$ pixels. All proposed methods must meet these computational constraints. No restrictions are imposed on memory footprint or inference time.

\paragraph{\textbf{Metric}}
For the ranking we adopted a scoring methodology inspired by the approach used in the NTIRE 2024 ESR and NTIRE 2025 ESR Challenges \cite{ren2024ninthntire2024efficient, ren2025tenthntire2025efficient}. Our formulation aims to aggregate multiple perceptual metrics into a single score relative to the baseline. For evaluation metrics where lower values indicate better performance, the score is computed as:

\[Score = \sum \lambda_i \cdot e^{(\frac{Metric^i}{Metric^i_{baseline}})}\]

Conversely, for metrics where higher values indicate better performance, the score is defined as:

\[Score = \sum \lambda_i \cdot e^{(\frac{Metric^i_{baseline}}{Metric^i})}\]

The evaluation metrics used are Perceptual Index (PI) \cite{Blau_2018_ECCV_Workshops}, CLIP Image Quality Assessment (CLIPIQA) \cite{wang2022exploringclipassessinglook}, and Multi-Dimension Attention Network for No-Reference Image Quality Assessment (MANIQA) \cite{yang2022maniqamultidimensionattentionnetwork}. Among commonly used metrics, we selected those exhibiting the highest Pearson Linear Correlation Coefficient (PLCC) and Spearman Rank-Order Correlation Coefficient (SRCC) \cite{su2025rethinkingimageevaluationsuperresolution}. The weighting coefficients were set as $\lambda_{PI} = 0.5$, $\lambda_{CLIPIQA} = 0.25$, and $\lambda_{MANIQA} = 0.25$, reflecting a balanced contribution between traditional non-deep learning NR-IQA metrics and deep learning-based metrics. All metrics were computed using the PIQA package\footnote{\href{https://github.com/francois-rozet/piqa}{https://github.com/francois-rozet/piqa}}.

While no standard quantitative measure for hallucinations or artifacts is available, a qualitative analysis will be conducted in a subsequent stage.

\paragraph{Training and validation phase}
The training phase lasted eight weeks. Due to Codabench’s lack of GPU support for NR-IQA metrics, participants were provided with the validation code to assess their progress locally. Validation was conducted on the RealSRSet and Real47 datasets.

\paragraph{Testing phase}
The test phase lasted one day. Participants were required to submit their code, a factsheet, and the output images for RealSRSet and Real47 to the organizers. The organizers verified the results by executing the submitted code under controlled conditions.

\section{Experimental Results}
\label{subsec:results}

We show in \Cref{tab:final-results} the initial results of the study. 

Besides the methods proposed in this paper, we included (i) BSRGAN \cite{zhang2021designingpracticaldegradationmodel} to assess the influence of training-time degradation modeling on test performance, and (ii) the top two entries from the NTIRE 2024 Efficient Super-Resolution Challenge \cite{ren2024ninthntire2024efficient, wan2024swiftparameterfreeattentionnetwork}—SPAN (XiaomiMM) and R2NET (Cao Group)—as strong PSNR-oriented, efficiency-focused baselines. 

\Cref{tab:final-results} reports both efficiency statistics (FLOPs and parameter counts) and perceptual quality indicators. Perceptual performance is evaluated from four complementary perspectives: PI (lower is better), CLIPIQA (higher is better), MANIQA (higher is better), and a scalar \emph{Score}, computed relatively to the Real-ESRGAN baseline (lower is better). This score is not a normalized metric, but rather a direct comparative measure of overall perceptual performance against the baseline.

\begin{table*}[t]
    \centering
    \caption{\textbf{Summary of results for EPSR}. The best and second-best results are highlighted in \textbf{bold} and \underline{underlined}, respectively. All metrics were obtained using the official evaluation code available at \href{https://github.com/brulonga/AIM-2025-Efficient-Perceptual-SR-Challenge}{https://github.com/brulonga/AIM-2025-EPSR-Challenge}.}
    \setlength\tabcolsep{2pt}
    \resizebox{0.9\textwidth}{!}{
        \begin{tabular}{l c c c c c c c}
            \toprule
            Team Name & Params$\downarrow$ (M) & FLOPs$\downarrow$ (G) & PI$\downarrow$ & CLIPIQA$\uparrow$ & MANIQA$\uparrow$ & Score & Rank \\\midrule
            Real-ESRGAN (baseline) & 16.6980 & 9293.9416 & 4.1442 & 0.5302 & 0.3283 & 2.7182 & - \\
            \midrule
            VPEG & 3.1684 & 1631.0842 & \textbf{3.1205} & \textbf{0.6544} & \textbf{0.3919} & \textbf{2.2015} & 1 \\
            MiAlgo & 3.5214 & 1987.3922 & \underline{3.7420} & \underline{0.5999} & \underline{0.3662} & \underline{2.4512} & 2 \\
            IPIU & 0.2762 & 132.1431 & 6.0676 & 0.3951 & 0.2722 & 3.9536 & 3 \\   
            \midrule
            BSRGAN & 16.6980 & 9293.9416 & 4.2112 & 0.5779 & 0.3350 & 2.6731 & -\\
            SPAN & 0.1507 & 77.7870 & 6.1198 & 0.3996 & 0.2748 & 3.9571 & - \\
            R2NET & 0.2148 & 103.2455 & 6.6837 & 0.3750 & 0.2811 & 4.3401 & - \\ 
           \bottomrule
        \end{tabular}
    }
    \label{tab:final-results}
\end{table*} 

\textbf{VPEG} achieved the highest overall performance, ranking first in all three perceptual metrics. Compared to Real-ESRGAN, it reduced PI by 24.7\%, increased CLIPIQA by 23.4\%, and increased MANIQA by 19.4\%, while using only $\sim$19.0\% of the parameters and $\sim$17.6\% of the FLOPs. These results demonstrate that substantial gains in perceptual quality can be achieved within a highly constrained efficiency budget.  

\textbf{MiAlgo} ranked second, delivering perceptual improvements comparable to VPEG: PI reduced by 9.7\%, CLIPIQA increased by 13.2\%, and MANIQA increased by 11.5\% over the baseline, with $\sim$21.1\% of the parameters and $\sim$21.4\% of the FLOPs. The final scores for VPEG and MiAlgo were 2.2015 and 2.4512, respectively, indicating closely matched performance.

The third-ranked method, \textbf{IPIU} (EFDN; winner of the NTIRE 2023 ESR Challenge \cite{Li_2023_CVPR}), is extremely lightweight ($\sim$1.65\% of the baseline parameters and $\sim$1.42\% of the FLOPs). However, as a distortion-oriented architecture primarily optimized for PSNR, its design is not fully aligned with the perceptual objectives of this track. We nonetheless consider it a relevant comparison point, as it illustrates the trade-off between distortion-focused optimization and perceptual quality under strict efficiency constraints.

Among the additional baselines, \textbf{SPAN} and \textbf{R2NET}~\cite{ren2025tenthntire2025efficient} show the high efficiency and PSNR performance characteristic of their original challenge context, but obtain comparatively low perceptual scores. This methods illustrate the extreme efficiency achievable in ESR. For efficient perceptual SR, this can be seen as a practical upper bound in FLOPs and parameter count, which is difficult to match while also maximizing perceptual quality.

Finally, \textbf{BSRGAN}~\cite{zhang2021designingpracticaldegradationmodel} slightly outperforms Real-ESRGAN on the PSR4K dataset, both in PI and CLIPIQA, while maintaining identical efficiency. This reinforces the importance of degradation modeling choices in determining perceptual outcomes, even when the network complexity remains constant.

\subsection{Extended Evaluation on Standard Perceptual SR Benchmarks}
\label{subsec:benchmarks}

To assess the generalization capabilities of the proposed and reference methods beyond the proposed PSR4K dataset, we evaluated all models on five widely adopted benchmarks for perceptual super-resolution. These datasets vary in content diversity, degradation characteristics, and difficulty, providing a comprehensive view of cross-dataset performance. Both perceptual quality metrics and runtime measurements are reported, enabling a joint analysis of visual fidelity and computational efficiency under diverse conditions.

As described in \Cref{subsec:datasets}, we tested all methods on the following datasets: DIV2K-LSDIR validation (\Cref{tab:DIV2K}), PIPAL validation (\Cref{tab:PIPAL}), RealSR validation (\Cref{tab:RealSR}), Real47 (\Cref{tab:Real47}), and RealSRSet (\Cref{tab:RealSRSet}).

Across all datasets, VPEG consistently achieves the best PI values, significantly improving over the Real-ESRGAN baseline. Specifically, VPEG reduces the PI by approximately 26.5\% and 30\% on the PIPAL and RealSR datasets, respectively, demonstrating substantial gains under real-world and GAN-based degradations. In contrast, MiAlgo leads in CLIPIQA and MANIQA (with the exception of RealSRSet), achieving improvements of roughly 34\% and 28\%, respectively, on the PIPAL dataset.

Traditional ESR methods generally underperform compared to perceptual SR solutions, particularly on PIPAL and RealSR, which contain more challenging degradations. These methods perform best on the DIV2K-LSDIR dataset, which features controlled bicubic downsampling (the degradation they were trained on). Among ESR methods, SPAN and EFDN show comparatively better results, surpassing R2NET.

\begin{table}[t]
    \centering
    \caption{Results on the DIV2K-LSDIR validation dataset.}
    \setlength\tabcolsep{2pt}
    \resizebox{\columnwidth}{!}{
        \begin{tabular}{l c c c c}
            \toprule
            \multicolumn{5}{c}{\textbf{DIV2K-LSDIR Validation Dataset}} \\
            \midrule
            Team Name & PI$\downarrow$ & CLIPIQA$\uparrow$ & MANIQA$\uparrow$ & Runtime\footnotemark[1]\ (ms) \\
            \midrule
            Real-ESRGAN (baseline) & 3.4401 & 0.5919 & 0.4082 & 118.6400 \\
            \midrule
            VPEG & \textbf{3.0813} & \underline{0.6426} & \underline{0.4273} & 35.7096 \\
            MiAlgo & \underline{3.3829} & \textbf{0.6790} & \textbf{0.4629} & 58.2874 \\
            IPIU & 5.3896 & 0.5000 & 0.3457 & 11.1416 \\
            \midrule
            BSRGAN & 3.5726 & 0.5963 & 0.4002 & 98.0291 \\
            SPAN & 5.4637 & 0.5054 & 0.3505 & \textbf{4.2781} \\
            R2NET & 6.1687 & 0.4935 & 0.3308 & \underline{5.1796} \\ 
            \bottomrule
        \end{tabular}
    }
    \label{tab:DIV2K}
\end{table}

\begin{table}[t]
    \centering
    \caption{Results on the PIPAL validation dataset.}
    \setlength\tabcolsep{2pt}
    \resizebox{\columnwidth}{!}{
        \begin{tabular}{l c c c c}
            \toprule
            \multicolumn{5}{c}{\textbf{PIPAL Validation Dataset}} \\
            \midrule
            Team Name & PI$\downarrow$ & CLIPIQA$\uparrow$ & MANIQA$\uparrow$ & Runtime\footnotemark[1]\ (ms) \\
            \midrule
            Real-ESRGAN (baseline) & 4.1254 & 0.4576 & 0.2783 & 69.0464 \\
            \midrule
            VPEG & \textbf{3.0366} & \textbf{0.6125} & \underline{0.3467} & 21.7737 \\
            MiAlgo & \underline{3.4911} & \underline{0.5885} &\textbf{ 0.3563} & 36.5912 \\
            IPIU & 6.5827 & 0.4199 & 0.2317 & 3.4071 \\
            \midrule
            BSRGAN & 3.8208 & 0.5154 & 0.2886 & 64.9269 \\
            SPAN & 6.5316 & 0.4019 & 0.2342 & \textbf{1.5343} \\
            R2NET & 6.7858 & 0.3806 & 0.2113 & \underline{1.8262} \\ 
            \bottomrule
        \end{tabular}
    }
    \label{tab:PIPAL}
\end{table}

\begin{table}[t]
    \centering
    \caption{Evaluation results were obtained on the RealSR validation dataset. Due to the large size of some images, certain images could not be processed. In total, only 58 images were successfully processed by all models; therefore, caution should be exercised when interpreting these results. It should be noted that the input images in this dataset are of a resolution ranging from 1K to 2K.}
    \setlength\tabcolsep{2pt}
    \resizebox{\columnwidth}{!}{
        \begin{tabular}{l c c c c}
            \toprule
            \multicolumn{5}{c}{\textbf{RealSR Validation Dataset}} \\
            \midrule
            Team Name & PI$\downarrow$ & CLIPIQA$\uparrow$ & MANIQA$\uparrow$ & Runtime\footnotemark[1]\ (ms) \\
            \midrule
            Real-ESRGAN (baseline) & 4.6645 & \underline{0.6479} & \underline{0.4050} & 10555.9394 \\
            \midrule
            VPEG & \textbf{3.2666} & 0.6115 & 0.3906 & 8638.6934 \\
            MiAlgo & \underline{4.3695} & \textbf{0.6932} & \textbf{0.4098} & 8619.2845 \\
            IPIU & 10.3502 & 0.5230 & 0.2974 & 8374.7927 \\
            \midrule
            BSRGAN & 5.7443 & 0.5288 & 0.3306 & 10465.1627 \\
            SPAN & 10.3741 & 0.5288 & 0.2983 & \underline{8269.8468} \\
            R2NET & 10.1132 & 0.4700 & 0.2990 & \textbf{8074.7931} \\ 
            \bottomrule
        \end{tabular}
    }
    \label{tab:RealSR}
\end{table}

\begin{table}[t]
    \centering
    \caption{Evaluation results on the Real47 dataset.}
    \setlength\tabcolsep{2pt}
    \resizebox{\columnwidth}{!}{
        \begin{tabular}{l c c c c}
            \toprule
            \multicolumn{5}{c}{\textbf{Real47 Dataset}} \\
            \midrule
            Team Name & PI$\downarrow$ & CLIPIQA$\uparrow$ & MANIQA$\uparrow$ & Runtime\footnotemark[1]\ (ms) \\
            \midrule
            Real-ESRGAN (baseline) & 3.5294 & 0.5999 & 0.3968 & 53.9262 \\
            \midrule
            VPEG & \textbf{3.0307} & \underline{0.6444} & \underline{}0.4107 & 17.2070 \\
            MiAlgo & \underline{3.4506} & \textbf{0.6771} & \textbf{0.4488} & 29.0713 \\
            IPIU & 5.6026 & 0.5315 & 0.2809 & 10.6158 \\
            \midrule
            BSRGAN & 3.5734 & 0.6042 & 0.3958 & 41.7053 \\
            SPAN & 5.6719 & 0.5233 & 0.2816 & \textbf{3.6206} \\
            R2NET & 6.2918 & 0.4599 & 0.3033 & \underline{7.0637} \\ 
            \bottomrule
        \end{tabular}
    }
    \label{tab:Real47}
\end{table}

\begin{table}[t]
    \centering
    \caption{Results on the RealSRSet dataset.}
    \setlength\tabcolsep{2pt}
    \resizebox{\columnwidth}{!}{
        \begin{tabular}{l c c c c}
            \toprule
            \multicolumn{5}{c}{\textbf{RealSRSet Dataset}} \\
            \midrule
            Team Name & PI$\downarrow$ & CLIPIQA$\uparrow$ & MANIQA$\uparrow$ & Runtime\footnotemark[1]\ (ms) \\
            \midrule
            Real-ESRGAN (baseline) & 4.8358 & 0.5875 & 0.3807 & 78.4476 \\
            \midrule
            VPEG & \textbf{4.0995} & \textbf{0.6635} & \textbf{0.4336} & 27.6264 \\
            MiAlgo & \underline{4.3723} & 0.6255 & \underline{0.4317} & 40.4732 \\
            IPIU & 6.0329 & 0.5397 & 0.3008 & 20.8097 \\
            \midrule
            BSRGAN & 4.6087 & \underline{0.6388} & 0.4110 & 51.1416 \\
            SPAN & 6.0452 & 0.5166 & 0.3025 & \textbf{3.9676} \\
            R2NET & 6.6692 & 0.5264 & 0.3027 & \underline{9.2148} \\ 
            \bottomrule
        \end{tabular}
    }
    \label{tab:RealSRSet}
\end{table}

\footnotetext[1]{The reported runtimes correspond to the average execution time obtained by running the provided evaluation code on an NVIDIA H100 80GB HBM3 GPU. Performance differences between Real-ESRGAN and BSRGAN can be attributed to variations in their original implementations (\href{https://github.com/XPixelGroup/BasicSR/blob/master/basicsr/archs/rrdbnet_arch.py}{Real-ESRGAN project} and \href{https://github.com/cszn/BSRGAN/blob/main/models/network_rrdbnet.py}{BSRGAN project}). 
}

For traditional perceptual SR methods such as Real-ESRGAN and BSRGAN, performance is largely consistent with expectations. Most datasets show similar results, except for PIPAL (where BSRGAN benefits from GAN-degradation training) and RealSR, where Real-ESRGAN demonstrates greater robustness to complex real-world degradations.

\begin{table}[t]
    \caption{Comparison of VPEG and MiAlgo performance relative to Real-ESRGAN metrics for each dataset, including the score differences between the two methods.}
    \centering
    \renewcommand{\arraystretch}{1.2} 
    \setlength{\tabcolsep}{3pt} 
    \footnotesize 
    \begin{tabular}{l c c c c c} 
        \toprule
        Team Name & DIV2K-LSDIR & PIPAL & RealSR & Real47 & RealSRSet \\
        \midrule
        VPEG & 2.5024 & 2.1294 & 2.4335 & 2.4712 & 2.3747 \\
        MiAlgo & 2.5383 & 2.2554 & 2.5840 & 2.5407 & 2.4783 \\
        Difference & 0.0359 & 0.1260 & 0.1505 & 0.0695 & 0.1036 \\
        \bottomrule
    \end{tabular}
    \label{tab:score}
\end{table}

\Cref{tab:score} reports a score relative to Real-ESRGAN for each dataset, providing a direct comparison between VPEG and MiAlgo. VPEG outperforms MiAlgo across all datasets, although the margin is smaller than on the PSR4K dataset. The average difference between VPEG and MiAlgo on these benchmarks is 0.0971, compared to 0.2497 on PSR4K. On DIV2K-LSDIR, the gap is minimal (0.0359). This analysis shows that both solutions achieve the largest improvement over Real-ESRGAN on the PIPAL dataset.

In terms of runtime performance, although these measurements should be interpreted with caution\footnotemark[1], VPEG consistently demonstrates significantly greater efficiency, requiring less than half the runtime of Real-ESRGAN across all evaluated datasets, with the exception of RealSR. This underscores the substantial improvement in computational efficiency.

\begin{table*}[t]
    \caption{Results obtained for each class in the PSR4K test set. The best class for each model is marked in blue and the worst in red.}
    \centering
    \renewcommand{\arraystretch}{1.2} 
    \setlength{\tabcolsep}{3pt} 
    \resizebox{\linewidth}{!}{
    \begin{tabular}{l c c c c c c c c c c c c c c c }
        \toprule
        Team Name & \multicolumn{3}{c}{Animals} & \multicolumn{3}{c}{Architecture} & \multicolumn{3}{c}{Art} & \multicolumn{3}{c}{Food} & \multicolumn{3}{c}{Nature} \\
        \cmidrule(lr){2-4} \cmidrule(lr){5-7} \cmidrule(lr){8-10} \cmidrule(lr){11-13} \cmidrule(lr){14-16} 
        & PI$\downarrow$ & CLIPIQA$\uparrow$ & MANIQA$\uparrow$ & PI$\downarrow$ & CLIPIQA$\uparrow$ & MANIQA$\uparrow$ & PI$\downarrow$ & CLIPIQA$\uparrow$ & MANIQA$\uparrow$ & PI$\downarrow$ & CLIPIQA$\uparrow$ & MANIQA$\uparrow$ & PI$\downarrow$ & CLIPIQA$\uparrow$ & MANIQA$\uparrow$ \\
        \midrule
        Real-ESRGAN & 4.1044 & 0.5387 & 0.3254 & \textcolor{blue}{3.4564} & \textcolor{blue}{0.5727} & \textcolor{blue}{0.3791} & 4.3428 & 0.5184 & 0.3009 & \textcolor{red}{4.9594} & \textcolor{red}{0.4307} & \textcolor{red}{0.2788} & 3.4804 & 0.5560 & 0.3139 \\
        \midrule
        VPEG & \textcolor{blue}{2.8712} & 0.6507 & 0.3635 & 3.1070 & 0.6550 & \textcolor{blue}{0.4342} & 3.1056 & \textcolor{blue}{0.6779} & 0.3852 & 3.4790 & \textcolor{red}{0.6187} & \textcolor{red}{0.3403} & 2.9538 & 0.6512 & 0.3702 \\
        MiAlgo & 3.5588 & 0.5981 & 0.3567 & 3.3224 & \textcolor{blue}{0.6604} & \textcolor{blue}{0.4324} & 3.9396 & 0.5614 & 0.3347 & \textcolor{red}{4.3602} & \textcolor{red}{0.4829} & \textcolor{red}{0.2874} & \textcolor{blue}{3.3214} & 0.6586 & 0.3707 \\
        IPIU & \textcolor{red}{6.3996} & \textcolor{blue}{0.4199} & 0.2923 & \textcolor{blue}{5.4120} & 0.3815 & \textcolor{blue}{0.2952} & 6.3712 & 0.3925 & 0.2690 & 6.1494 & 0.3894 & \textcolor{red}{0.2369} & 5.7130 & 0.4077 & 0.2850 \\
        \midrule
        BSRGAN & 4.1098 & \textcolor{blue}{0.5993} & 0.3199 & 3.7090 & 0.5750 & \textcolor{blue}{0.3856} & 4.4004 & 0.5882 & 0.3132 & 4.5946 & 0.5495 & \textcolor{red}{0.2912} & \textcolor{blue}{3.6012} & 0.5904 & 0.3101 \\
        SPAN & 6.3950 & 0.4180 & 0.2952 & \textcolor{blue}{5.4864} & 0.3943 & \textcolor{blue}{0.3002} & 6.3970 & 0.3910 & 0.2703 & 6.2188 & 0.3908 & 0.2380 & 5.7620 & 0.4040 & 0.2864 \\
        R2NET & 6.9088 & 0.3869 & 0.3033 & \textcolor{blue}{6.1674} & 0.3935 & \textcolor{blue}{0.3098} & 6.8458 & 0.3638 & 0.2682 & 6.8612 & 0.3361 & \textcolor{red}{0.2476} & 6.5912 & 0.3761 & 0.2855 \\
        \midrule
         & \multicolumn{3}{c}{Objects} & \multicolumn{3}{c}{Portraits} & \multicolumn{3}{c}{Sports} & \multicolumn{3}{c}{Text} & \multicolumn{3}{c}{Urban} \\
        \cmidrule(lr){2-4} \cmidrule(lr){5-7} \cmidrule(lr){8-10} \cmidrule(lr){11-13} \cmidrule(lr){14-16} 
        & PI$\downarrow$ & CLIPIQA$\uparrow$ & MANIQA$\uparrow$ & PI$\downarrow$ & CLIPIQA$\uparrow$ & MANIQA$\uparrow$ & PI$\downarrow$ & CLIPIQA$\uparrow$ & MANIQA$\uparrow$ & PI$\downarrow$ & CLIPIQA$\uparrow$ & MANIQA$\uparrow$ & PI$\downarrow$ & CLIPIQA$\uparrow$ & MANIQA$\uparrow$ \\
        \midrule
        Real-ESRGAN & 3.9726 & 0.5359 & 0.3353 & 4.3434 & 0.5189 & 0.3135 & 4.6794 & 0.5310 & 0.3252 & 4.5560 & 0.5576 & 0.3610 & 3.5492 & 0.5423 & 0.3495 \\
        \midrule
        VPEG & 3.0282 & 0.6523 & 0.4038 & 3.0018 & 0.6726 & 0.3774 & 3.1226 & 0.6665 & 0.3931 & \textcolor{red}{3.6368} & 0.6566 & 0.4300 & 2.9012 & 0.6425 & 0.4221 \\
        MiAlgo & 3.6186 & 0.6190 & 0.3786 & 3.7072 & 0.5855 & 0.3431 & 4.1018 & 0.5743 & 0.3513 & 4.2674 & 0.6326 & 0.4019 & 3.2238 & 0.6262 & 0.4051 \\
        IPIU & 5.9424 & 0.4136 & 0.2724 & 6.2422 & 0.4062 & 0.2662 & 6.5130 & 0.3974 & 0.2570 & 6.2766 & 0.4132 & 0.2886 & 5.6584 & \textcolor{red}{0.3296} & 0.2593 \\
        \midrule
        BSRGAN & 4.0920 & 0.5668 & 0.3415 & 4.2946 & 0.5970 & 0.3256 & \textcolor{red}{4.7294} & 0.5897 & 0.3397 & 4.8254 & 0.5795 & 0.3674 & 3.7566 & \textcolor{red}{0.5440} & 0.3566 \\
        SPAN & 6.0010 & 0.4176 & 0.2748 & 6.3114 & 0.4147 & 0.2690 & \textcolor{red}{6.5666} & 0.4007 & 0.2599 & 6.3116 & \textcolor{blue}{0.4257} & 0.2906 & 5.7486 & \textcolor{red}{0.3392} & 0.2633 \\
        R2NET &  6.5486 & 0.3880 & 0.2831 & 6.8230 & 0.3829 & 0.2831 & \textcolor{red}{6.9994} & 0.3692 & 0.2659 & 6.7734 & \textcolor{blue}{0.4265} & 0.2996 & 6.3162 & \textcolor{red}{0.3268} & 0.2646 \\
        \bottomrule
    \end{tabular}}
    \label{tab:classes}
\end{table*}

\begin{table*}[t]
    \caption{The mean, median, and standard deviation were computed for the set of metric values obtained for each class in the PSR4K test set. The best overall results per class are highlighted in blue, and the worst in red.}
    \centering
    \renewcommand{\arraystretch}{1.2} 
    \setlength{\tabcolsep}{3pt} 
    \resizebox{\linewidth}{!}{
    \begin{tabular}{l c c c c c c c c c c c c c c c }
        \toprule
        Team Name & \multicolumn{3}{c}{PI$\downarrow$} & \multicolumn{3}{c}{CLIPIQA$\uparrow$} & \multicolumn{3}{c}{MANIQA$\uparrow$ } \\
        \cmidrule(lr){2-4} \cmidrule(lr){5-7} \cmidrule(lr){8-10}
        & Mean & Median & Standard Deviation & Mean & Median & Standard Deviation & Mean & Median & Standard Deviation \\
        \midrule
        Real-ESRGAN  & 4.1444 & 4.2236 & \textcolor{red}{0.5269} & 0.5302 & 0.5373 & 0.0389 & 0.3283 & 0.3253 & 0.0294 \\
        \midrule
        VPEG         & \textcolor{blue}{3.1207} & \textcolor{blue}{3.0669} & \textcolor{blue}{0.2486} & \textcolor{blue}{0.6544} & \textcolor{blue}{0.6537} & \textcolor{blue}{0.0166} & \textcolor{blue}{0.3920} & \textcolor{blue}{0.3891} & 0.0307 \\
        MiAlgo       & 3.7421 & 3.6629 & 0.4080 & 0.5999 & 0.6085 & \textcolor{red}{0.0530} & 0.3662 & 0.3637 & \textcolor{red}{0.0414} \\
        IPIU         & 6.0678 & 6.1958 & 0.3682 & 0.3951 & 0.4018 & 0.0260 & \textcolor{red}{0.2722} & \textcolor{red}{0.2707} & \textcolor{blue}{0.0184} \\
        \midrule
        BSRGAN       & 4.2113 & 4.2022 & 0.4335 & 0.5779 & 0.5838 & 0.0192 & 0.3351 & 0.3327 & 0.0288 \\
        SPAN         & 6.1198 & 6.2651 & 0.3522 & 0.3996 & 0.4023 & 0.0245 & 0.2748 & 0.2725 & 0.0189 \\
        R2NET        & \textcolor{red}{6.6835} & \textcolor{red}{6.7982} & 0.2716 & \textcolor{red}{0.3750} & \textcolor{red}{0.3795} & 0.0286 & 0.2811 & 0.2831 & 0.0197 \\ 
        \bottomrule
    \end{tabular}}
    \label{tab:statistics}
\end{table*}

\subsection{Per-Class Performance on the PSR4K Dataset}
\label{subsec:class}

The PSR4K dataset comprises ten distinct semantic categories. Analyzing performance at the class level provides valuable insights into the strengths and limitations of each method. Class-wise evaluation reveals perceptual quality patterns that may be obscured in aggregated metrics, such as performance degradation in texture rich scenes or improvements in structured environments.

As shown in Table \ref{tab:classes}, several consistent trends emerge. The architecture category is the most favorable across all methods, likely due to the structured nature of these scenes and their prevalence in training datasets such as DIV2K, FLICKR2K, and LSDIR. Similarly, the animals and nature categories exhibit above-average performance, which may also be attributed to their frequent appearance in training data.

In contrast, the food category consistently yields the lowest scores. This under-performance is likely due to the rich textures and fine details of food imagery, combined with its under-representation in existing datasets. Urban and sports scenes also pose challenges, particularly for PSNR-oriented methods. Interestingly, metric correlations within these categories are less consistent, with some metrics favoring certain methods and classes while others penalize them.

Surprisingly, most models perform reasonably well on the text category. While PI tends to penalize blurry or imprecise lettering, CLIPIQA and MANIQA achieve above-average scores, comparable to those in the architecture category. This suggests that these metrics prioritize global perceptual quality over fine-grained textual details.

The art, portraits, and objects categories tend to align closely with the overall average performance, showing neither significant advantage nor disadvantage.

In summary, the poor performance on food images can be attributed to both their complex textures and lack of representation in training datasets. Meanwhile, categories such as sports, urban scenes, and text present more intrinsic challenges for perceptual super-resolution, as they expose inconsistencies in metric behavior and highlight the limitations of current evaluation frameworks.

In addition to reporting the metric values per class, we computed descriptive statistics (mean, median, and standard deviation) across the set of metrics obtained for each semantic category, as shown in \Cref{tab:statistics}. This analysis provides insight into the stability of model performance under varying content conditions. A lower standard deviation reflects greater robustness to scene variability, whereas higher variability may indicate sensitivity to specific visual patterns.

Once again, the VPEG solution emerges as the most robust method. It achieves a standard deviation of 0.2486 for the PI metric, significantly outperforming Real-ESRGAN (0.5269). Similarly, for CLIPIQA, VPEG attains a standard deviation of 0.0166, compared to 0.0389 for Real-ESRGAN. Although VPEG's standard deviation on the MANIQA metric is slightly higher than that of Real-ESRGAN and BSRGAN, it remains competitive.

Conversely, while the MiAlgo team delivers results comparable to VPEG in overall performance, their model exhibits greater variability across categories. This suggests reduced robustness and increased sensitivity to diverse textures and semantic content. Overall, this analysis further reinforces the strength of the VPEG solution, not only as a top‑performing approach, but also as a robust alternative that surpasses traditional methods such as Real‑ESRGAN and BSRGAN

\begin{figure*}[t]
    \centering
    \includegraphics[width=\linewidth]{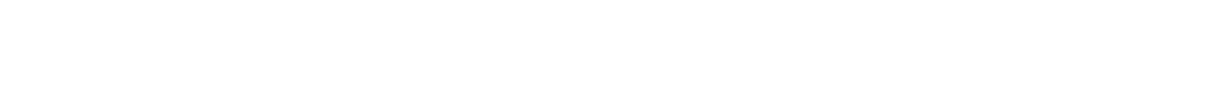}\par
    \includegraphics[width=\linewidth]{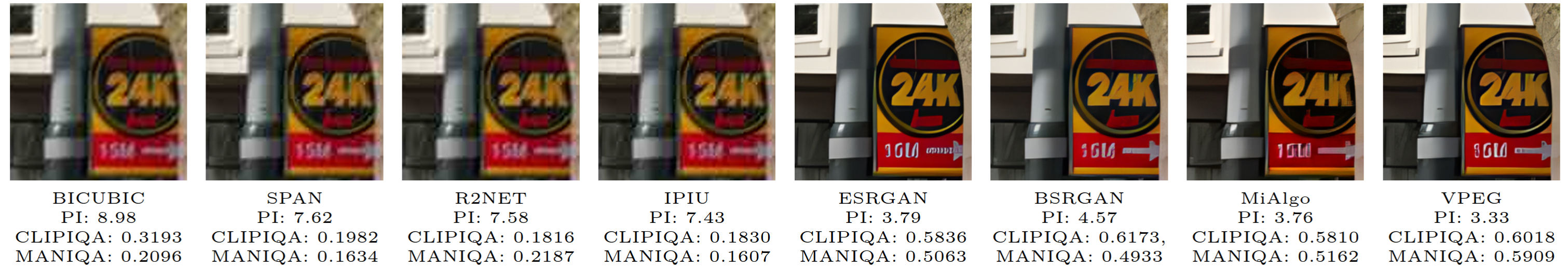}\par
    \includegraphics[width=\linewidth]{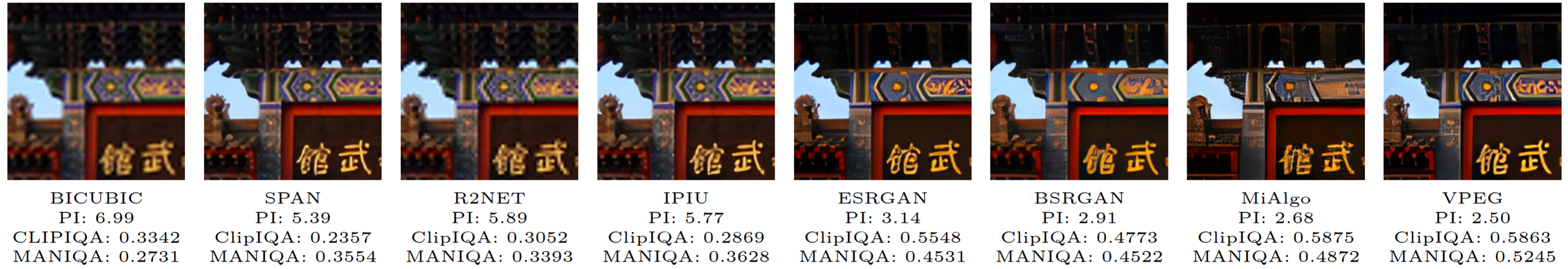}\par
    \includegraphics[width=\linewidth]{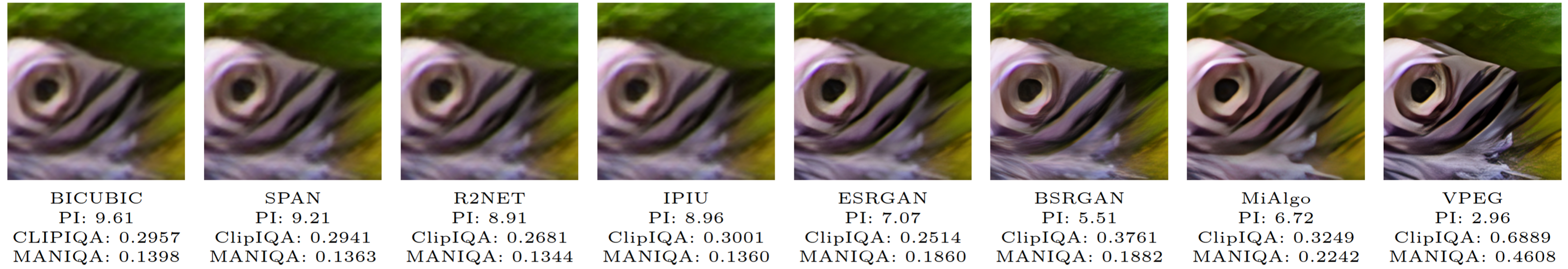}\par
    \includegraphics[width=\linewidth]{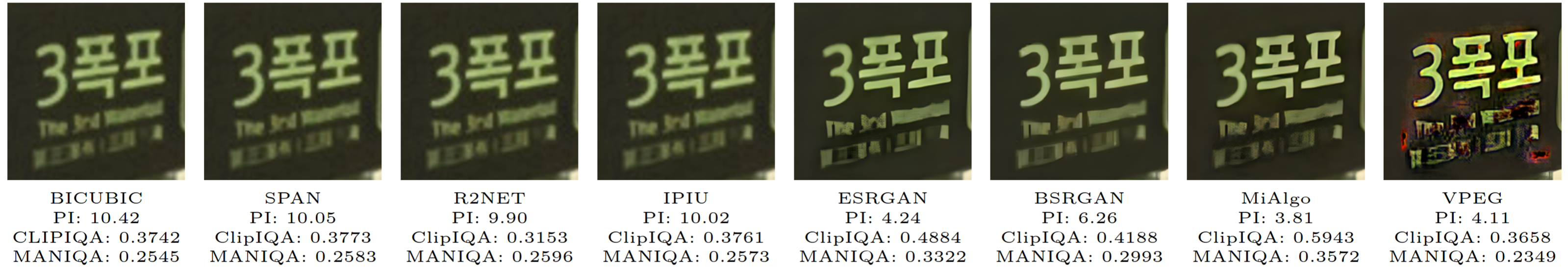}\par
    \includegraphics[width=\linewidth]{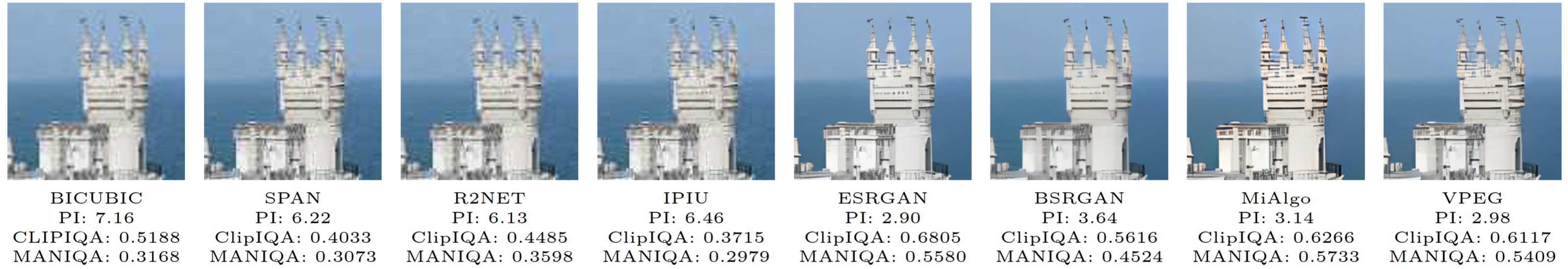}\par
    \includegraphics[width=\linewidth]{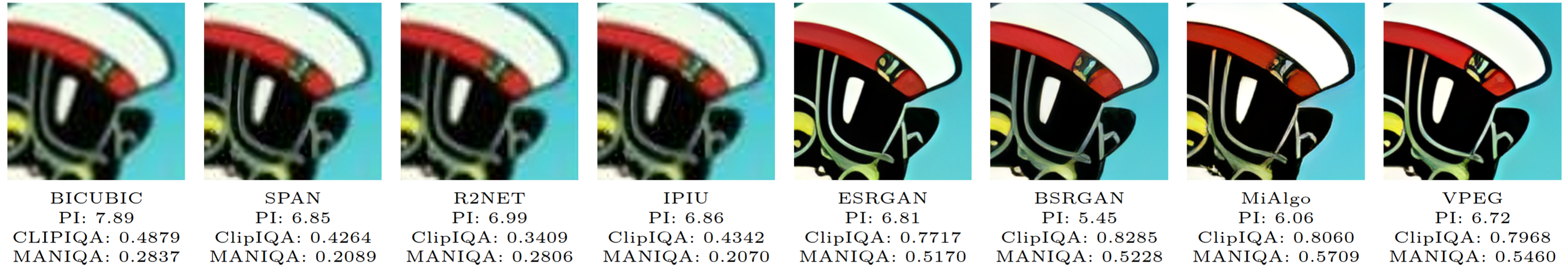}
    \caption{\textbf{Qualitative comparison of super-resolution results across multiple datasets.} 
    For each dataset, cropped image regions are arranged from left to right in approximate order of increasing perceptual quality, as indicated by the corresponding quantitative metrics. The corresponding datasets for each crop, from top to bottom, are: PSR4K, DIV2K-LSDIR validation set, RealSR validation set, PIPAL validation set, Real47, and RealSRSet. Images have been downscaled from their original upscaled resolution and compressed to meet compilation constraints, which may slightly affect their visual fidelity. Although the image displays 'ESRGAN', we are in fact referring to 'Real-ESRGAN'; the omission of 'Real' is purely for aesthetic purposes.}
    \label{fig:qualitative-results}
\end{figure*}

\subsection{Qualitative Comparison Across Benchmarks}

To complement the quantitative results, we present a visual comparison of representative image crops from each benchmark, processed by all evaluated models. Each crop is accompanied by its corresponding perceptual metric scores, enabling a direct correlation between numerical performance and perceived image quality. This comparison is illustrated in \Cref{fig:qualitative-results}.

It is important to note that the images have been rescaled and, in some cases, compressed for visualization purposes within the figure. As a result, certain visual enhancements introduced by the models may be partially lost. With this in mind, it becomes evident that non-perceptual methods, such as SPAN, R2NET, and EFDN, fail to deliver meaningful perceptual improvements. These models tend to produce results that are only marginally better than bicubic upsampling in terms of PI, while CLIPIQA and MANIQA scores often deteriorate. In many cases, these methods do not clearly surpass bicubic interpolation, which should be considered a baseline requirement.

In contrast, perceptual methods demonstrate clear qualitative improvements. Models such as Real-ESRGAN, BSRGAN, MiAlgo, and VPEG not only outperform bicubic upsampling visually, but also achieve significantly better scores across perceptual metrics. The VPEG solution continues to stand out as the top-performing approach, closely followed by MiAlgo, with Real-ESRGAN and BSRGAN occasionally surpassing them in specific samples (Real47 and RealSRSet samples, respectively).

However, this qualitative comparison also underscores a critical limitation of current perceptual metrics: their inability to effectively penalize hallucinations or artifacts. As observed in the RealSR and Real47 crops, both VPEG and MiAlgo introduce noticeable artifacts. In particular, the VPEG solution severely degrades the RealSR crop, producing results that are clearly flawed to the human eye. Yet, despite these issues, the perceptual metrics fail to reflect the degradation, assigning top-performing scores to these outputs. In contrast, traditional models such as Real-ESRGAN and BSRGAN appear more robust in these cases, with Real-ESRGAN delivering an exceptionally accurate reconstruction in the Real47 sample, free of hallucinations (or artifacts) and accompanied by the highest metric scores.

It is also worth highlighting that the VPEG and MiAlgo solutions excel in the PIPAL dataset, producing high-quality reconstructions that align well with perceptual metrics. This observation supports the analysis presented in \Cref{subsec:benchmarks}, where PIPAL was identified as the benchmark most responsive to perceptual improvements.

\section{Methods}
\label{sec:methods}

In the following section, we outline each contributor’s solution, all designed to satisfy our efficiency constraints, while maximizing perceptual super‑resolution performance. 

Note that the method descriptions were provided by each team as their contribution to this report.

\subsection{VPEG}
\label{subsec:VPEG}

\begin{center}

\vspace{2mm}
\noindent\emph{\textbf{Spatially-Adaptive Feature Modulation for Efficient Perceptual Image Super-Resolution}}
\vspace{2mm}

\noindent\emph{Ke Wu, Long Sun, Lingshun Kong, Jinshan Pan, Jiangxin Dong, Jinhui Tang}

\vspace{2mm}

\noindent\emph{Nanjing University of Science and Technology}



\end{center}


\paragraph{Method}

The VPEG team uses SAFMN architecture~\cite{sun2023safmn} as the baseline model in their work. The original SAFMN open-source model exceeded the challenge efficiency constraints using 2888.23~GFLOPs.

To meet efficiency requirements without significantly compromising quality, the VPEG team proposed a reduced-complexity variant, SAFMN-L, which maintained 16 blocks but reduced the channel dimension from 128 to 96. The overall architecture is shown in \Cref{fig:team_VPEG}. They successfully created a perceptual version of SAFMN by incorporating Perceptual~\cite{johnson2016perceptuallossesrealtimestyle}, LDL~\cite{jie2022LDL}, GAN~\cite{wang2021realesrgantrainingrealworldblind} and AESOP losses~\cite{lee2025auto}. No pre-trained SR weights were used for fine-tuning; however, the AESOP pre-trained autoencoder was employed within the loss computation.

\paragraph{Training Details} A three-stage training strategy was employed to progressively enhance performance:

\begin{enumerate} 

\item Stage~I: SAFMN-L was trained with 192$\times$192 input patches, a batch size of 64, an initial learning rate of $3\times10^{-4}$ decayed to $1\times10^{-6}$ via cosine annealing, using a weighted combination of L1 loss (1.0) and FFT-based L1 loss (0.05) for 300k iterations using Adam.

\item Stage~II: Used the same patch size, a batch size of 36, a learning rate schedule of $1\times10^{-4}$ to $1\times10^{-6}$, and minimized L1 (1.0), Perceptual (0.1), LDL (1.0), and GAN (0.1) losses over 300k iterations. 

\item Stage~III: Retained the patch size, used a batch size of 16, and applied the same learning rate schedule while optimizing a combination of AESOP (1.0), Perceptual (0.1), LDL (1.0), and GAN (0.1) losses for 100k iterations. 
\end{enumerate}

Perceptual loss was defined using pre-activation {conv1–conv5} feature maps from VGG19 (weights {0.1, 0.1, 1, 1, 1}). GAN loss employed a Spectral UNet discriminator, optimized with a CosineAnnealing scheduler (min LR 1e-6). EMA strategy was applied throughout training.

All experiments were conducted with PyTorch on NVIDIA RTX~3090 GPUs, with a memory footprint of approximately 20–23~GB during training. Data preprocessing, augmentation, and training procedures followed BasicSR~\cite{basicsr}, and the total training duration was about eight days. The training datasets used were DIV2K and LSDIR, and degraded images were obtained following the Real-ESRGAN degradation pipeline.

\subsection{MiAlgo}
\label{subsec:MiAlgo}

\begin{center}

\vspace{2mm}
\noindent\emph{\textbf{TinyESRGAN: A Lightweight ESRGAN Variant for Real-World Image Super-Resolution}}
\vspace{2mm}

\noindent\emph{Tianyu Hao, Yuxuan Qiu, Yueqi Yang, \\Chaoyu Feng, Na Jiang, Dongqing Zou, Lei Lei}

\vspace{2mm}

\noindent\emph{Xiaomi Inc.\\Capital Normal University}



\end{center}


\paragraph{Method}
The overall architecture of the MiAlgo team’s solution, illustrated in \Cref{fig:mialgo}, is based on the ESRGAN framework~\cite{wang2018esrganenhancedsuperresolutiongenerative} and redesigned as a lightweight variant, named TinyESRGAN, to achieve efficient 4× image super-resolution. Several structural modifications were introduced by the MiAlgo team to reduce computational complexity while maintaining comparable perceptual performance to the baseline. Specifically, the number of Residual-in-Residual Dense Blocks (RRDBs) was reduced to 17, the number of intermediate feature channels in each RRDB was set to 32, and the channel growth rate within each dense block was set to 18. These adjustments resulted in an approximate 79\% reduction in computational cost relative to the original ESRGAN architecture, thereby improving its suitability for deployment on resource-constrained platforms such as mobile and embedded devices.

\paragraph{Training Details}
The TinyESRGAN model was implemented in the PyTorch framework and trained on a single NVIDIA H20 GPU following a multi-stage strategy:
\begin{enumerate}
    \item \textbf{Stage~I:} Optimization with a combination of MSE loss (1.0) and LPIPS~\cite{zhang2018perceptual} perceptual loss (1.0) over 500{,}000 iterations, with a batch size of~32 and an initial learning rate of $3\times10^{-4}$, decayed via cosine annealing.
    \item \textbf{Stage~II:} Addition of a GAN loss (0.1)~\cite{wang2021realesrgantrainingrealworldblind}, with training continuing for an additional 250{,}000 iterations, keeping the batch size unchanged and reducing the learning rate to $1\times10^{-4}$.
\end{enumerate}

The network was trained for 4× super-resolution, mapping 128×128 low-resolution inputs to 512×512 high-resolution outputs. Adam was used for optimization in both stages. For GAN loss, the MiAlgo team followed the Real-ESRGAN setup with a Spectral-UNet discriminator, but without applying a learning-rate scheduler. Training employed high-resolution images from DIV2K, Flickr2K, and OST as ground truth, with realistic low-resolution counterparts generated via the Real-ESRGAN degradation pipeline. Training procedures relied on BasicSR \cite{basicsr}, and EMA was applied throughout.

\subsection{IPIU}
\label{subsec:IPIU}

\begin{center}

\vspace{2mm}
\noindent\emph{\textbf{Data Augmented Edge Distillation for Resource Efficient Image Super-Resolution}}
\vspace{2mm}

\noindent\emph{Lianping Lu, Heng Yang, Meilin Gao}

\vspace{2mm}

\noindent\emph{Intelligent Perception and Image Understanding Lab, Xidian University}

\vspace{2mm}


\end{center}


\paragraph{Method}
The proposed solution is built upon the Edge-enhanced Feature Distillation Network (EFDN)~\cite{wang2022edge}, a lightweight yet high-performing super-resolution model that combines block design, neural architecture search, and tailored loss functions to achieve an optimal balance between reconstruction quality and computational efficiency. The architecture of the EFDN is presented in \Cref{fig:efdn}. At its core, EFDN employs an Edge-Enhanced Diverse Branch Block (EDBB), which consolidates and extends existing re-parameterization techniques into a versatile, multi-branch module that enhances both structural and high-frequency edge feature extraction \cite{ding2019acnetstrengtheningkernelskeletons, ding2021diversebranchblockbuilding, zhang2021edge}. Multiple reparameterizable branches capture complementary edge and texture cues, which are then fused into a standard convolution to preserve inference efficiency. 

\paragraph{Training Details}
The model was implemented in PyTorch~\cite{paszke2019pytorch} and trained on a single NVIDIA RTX~3090 GPU using the Flickr2K dataset~\cite{lim2017enhanced} as ground truth. For this challenge training was performed for 15 hours with a batch size of 64, optimizing an L1 loss with the Adam optimizer~\cite{adam2014method} and an initial learning rate of $1\times10^{-3}$, decayed using a cosine annealing schedule. Horizontal and vertical flipping were applied for data augmentation.


\begin{center}
\begin{minipage}{0.95\textwidth}
\centering

\includegraphics[width=\textwidth]{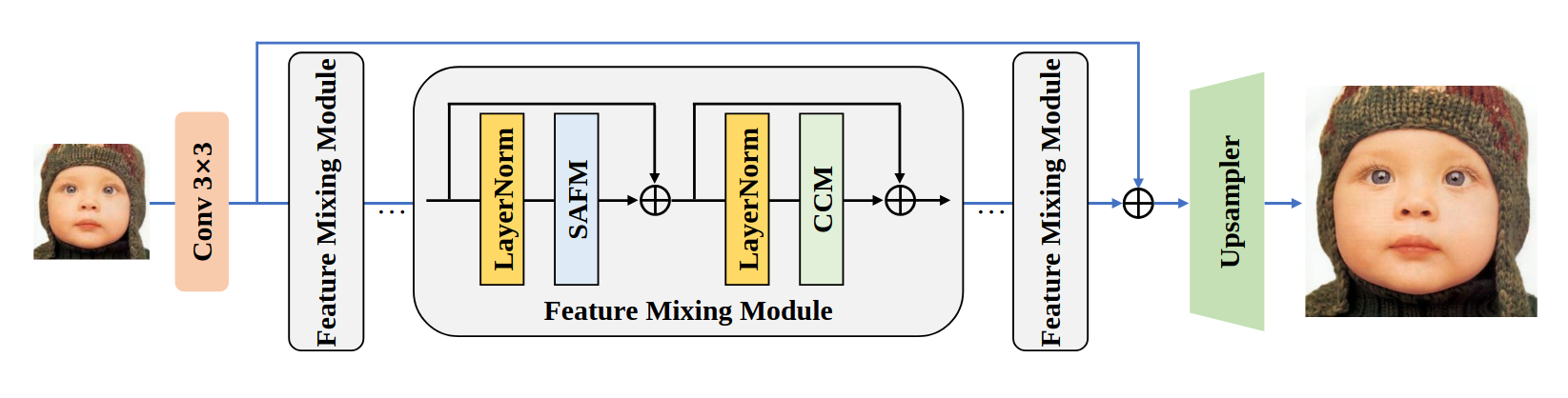}
\captionof{figure}{An overview of the proposed SAFMN-L in \Cref{subsec:VPEG} by the VPEG team. SAFMN employs a series of feature mixing modules (FMMs) to process deep-level features. The FMM block is composed of a spatially-adaptive feature modulation (SAFM) and a convolutional channel mixer (CCM).}
\label{fig:team_VPEG}

\vspace{1cm}

\includegraphics[width=\textwidth]{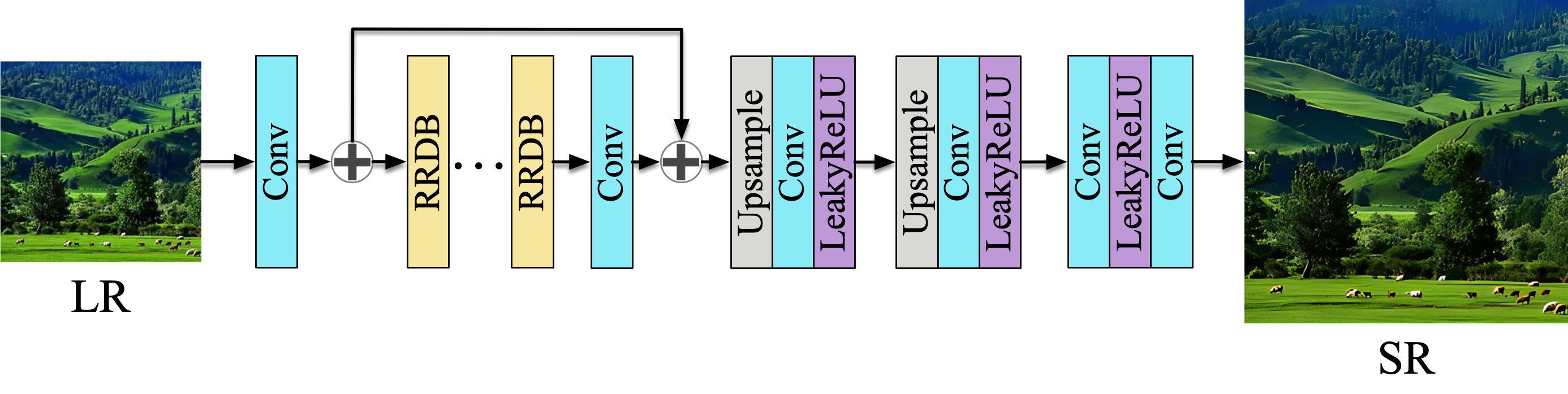}
\captionof{figure}{Overview of TinyESRGAN proposed in \Cref{subsec:MiAlgo} by MiAlgo team.}
\label{fig:mialgo}

\vspace{1cm}

\includegraphics[width=\textwidth]{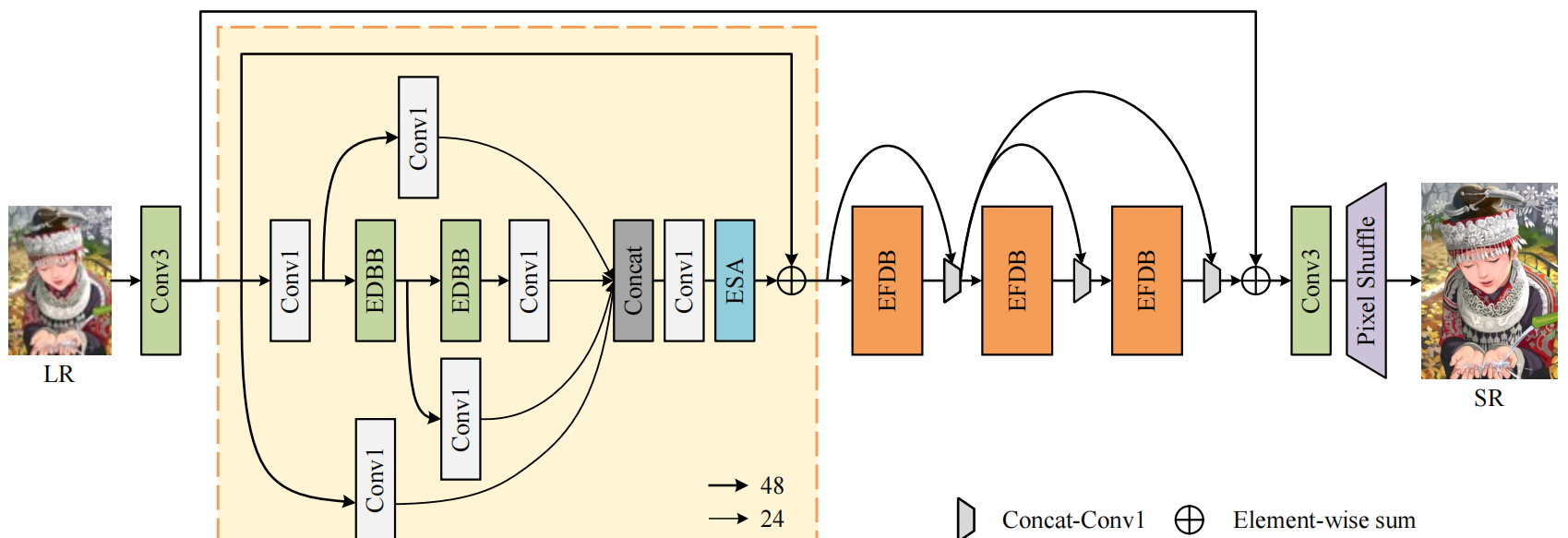}
\captionof{figure}{Overview of EFDN architecture presented by IPIU team in \Cref{subsec:IPIU}.}
\label{fig:efdn}

\end{minipage}
\end{center}

\clearpage
\section{Conclusions}
\label{sec:conclusion}

We conclude the following points from this study and the proposed benchmarks:

\begin{itemize}

    \item The challenge introduced a new test set, \textbf{PSR4K}, divided into ten semantic categories to facilitate more fine-grained analysis. This dataset has proven valuable for in‑depth research and, as its name suggests, consists of 4K‑resolution images, aiming to establish a benchmark for 4K SR.
    
    \item The results corroborate that improving perceptual image quality while adhering to strict efficiency constraints is indeed possible, thereby opening the door to a relatively unexplored research direction.
    
    \item Despite improvements in perceptual quality metrics, the studied methods tend to produce visual artifacts, which raises questions about a possible efficiency-perception trade-off. Moreover, this emphasizes the need for new perceptual quality metrics that remain robust in the presence of artifacts or hallucinations.
    
    \item Notably, several widely used techniques for efficient super‑resolution, such as knowledge distillation~\cite{Huang2020RealTimeIF}, re‑parameterization (applied only by the IPIU team)~\cite{ding2019acnetstrengtheningkernelskeletons, ding2021diversebranchblockbuilding, zhang2021edge}, and pruning~\cite{liu2017learningefficientconvolutionalnetworks, molchanov2017pruningconvolutionalneuralnetworks}, were not employed in this analysis, thereby leaving untapped potential for future exploration. 
\end{itemize}

\section*{Acknowledgments}
This work was partially supported by the Alexander von Humboldt Foundation. We thank the AIM 2025 sponsors: AI Witchlabs and University of W\"urzburg (Computer Vision Lab).

Cidaut AI thank Supercomputing of Castile and Leon (SCAYLE. Leon, Spain) for assistance with
the model training and GPU resources.

{
    \small
    \bibliographystyle{ieeenat_fullname}
    \bibliography{main}
}

\end{document}